\documentclass{llncs}

\usepackage{color}
\usepackage{tikz}
\usepackage{pgfplots}
\usepackage{url}
\usepackage{booktabs}

\usepackage{pgfplots}
\usetikzlibrary{pgfplots.groupplots}
\usetikzlibrary{matrix,calc}

\pgfplotsset{compat=1.3}

\newcommand{\negspace}{\vspace{-1.9mm}}

\long\def\ignore#1{}

\newcommand{\tempclearpage}[0]{}
\renewcommand{\tempclearpage}[0]{\clearpage}
\newcommand{\mysection}[1]{\tempclearpage\section{#1}}

\let\AJPii=\ignore

\def\makeclean{
	\let\AJP=\ignore 
	\let\DK=\ignore 
	\let\TS=\ignore 
    \let\tempclearpage=\relax
    \let\mysection=\section
}

\title{
Algorithm Configuration: Learning policies for the quick termination of poor performers
}

\author{
Daniel Karapetyan\inst{1} 
\and
Andrew J. Parkes\inst{2}
\and
Thomas St\"utzle\inst{3}
}

\institute{
Institute for Analytics and Data Science, University of Essex, UK
\and
ASAP Research Group, School of Computer Science, University of Nottingham, UK
\and
IRIDIA, CoDE, Université Libre de Bruxelles (ULB), Brussels, Belgium
}

\begin{document}


\maketitle

\negspace
\begin{abstract}
 One way to speed up the algorithm configuration task is to use short runs instead of long  runs as much as possible, but without discarding the configurations that eventually do well on the long runs.
 We consider the problem of selecting the top performing configurations of the Conditional Markov Chain Search (CMCS), a general algorithm schema that includes, for examples, VNS\@.
 We investigate how the structure of performance on short tests links with those on long tests, showing that significant differences arise between test domains. 
 We propose a ``performance envelope'' method to exploit the links; that learns when runs should be terminated, but that automatically adapts to the domain.
\negspace
\end{abstract}

\negspace
\negspace

\section{Introduction}

 Careful configuration of algorithms can lead to a significant improvement in performance~\cite[and many others]{HutterEtal2011:SMAC}.
 This is usually done by searching in the space of configurations and evaluating each configuration on a set of target instances.
 However, such instances are often large and will require long runs, so direct and complete usage of such intended instances problems will be overly time-consuming.  
 A natural desire is that, in a justified fashion, we should be able to reduce the run times by exploiting the results of ``short runs'' in order to configure for ``long runs'': There is a need to learn how to extrapolate from ``short'' to ``long''.
 This suggests that machine learning 
 methods should be applied to collections of such ``short-run data'', to analyse patterns, and so produce predictions for the performance in the long runs.
This view suggests that for algorithm configuration at least  3 different `generic' spaces are relevant:
\begin{itemize}
\item C ``Configuration space'' -- the direct parameters of the algorithms.
\item S  ``Short-run space'' -- the space of (detailed) results using short runs. \AJPii{A combination of small instances with relatively long runs, and short runs on larger instances.}
\item L  ``Long-run space'' -- results from long runs.
\end{itemize}
 A common procedure would be to do a (local) search in C-space, using fitnesses obtained from L-space. 
 In this context, the usage of machine learning might be to develop a mapping from the C-space to the L-space, e.g.\ see SMAC~\cite{HutterEtal2011:SMAC}.
%
 In this paper, we instead study the potential for learning the mappings from S-space to L-space -- aiming to exploit how short runs are able to predict longer term behaviours.
 The goal is to use such information to optimise the policies for when a trial of a particular configuration should be terminated -- because there is high confidence it will not lead to a good final solution.

\section{Experimental Setup}
\negspace

 We used CMCS, a recent framework that defines the behaviour of a multi-component optimisation algorithm with a set of numeric parameters~\cite{SPLP,BBQP}.
 We used three problem domains: the Simple Plant Location Problem (SPLP)~\cite{SPLP}, the Far From Most String Problem (FFMSP) (the details of our components, testbed, etc.\ are not yet published) and the Bipartite Boolean Quadratic Problem (BBQP)~\cite{BBQP}.
 We generated all `meaningful' 3-component configurations with deterministic control mechanism, thus ending up with a finite number of configurations. 
 We do not include details of exactly what these mean (see \cite{BBQP,SPLP}), as for the purposes of this paper, these can simply be regarded as a categorical set of potential options, and defining the ``C-space''.
 The goal of the work is to find configurations that are the best performers, with respect to the long-run L-space, but exploiting their properties with respect to the short-run S-space in order to reduce the overall time budget.

 The full time budget for each `long' run was selected as 1024ms.
 Ten random instances were generated for each domain, with the size chosen to make them hard enough for this time budget.
 We then generated performance data\footnote{A URL will be provided} for each domain, by solving each of the ten instances by each of the configurations, and recording the solution quality at 1ms, 2ms, 4ms, \ldots, 1024ms.
 Objective values were scaled to $[0, 1]$ for each instance, and then averaged over the data instances. 
 Hence, quality 0 (resp.\ 1) means that, for each instance, the configuration yielded solution as good (resp.\ bad) as any other configuration within the full 1024ms time budget.

\tikzset{
	top conf/.style={black, ultra thick},
	cutoff 1/.style={black, ultra thick, dash pattern=on 5pt off 2pt},
	cutoff 5/.style={black, ultra thick, dotted},
}

\begin{figure}[tb]
\begin{tikzpicture}
\begin{groupplot}[
	group style = {
    	group size = 2 by 2, 
        horizontal sep = 50pt,
        vertical sep = 60pt
    }, 
    width = 6.0cm, 
    height = 4.2cm, 
    grid=major,
  	xlabel={Time, s},
]
\nextgroupplot[ 
    legend style = {
    	column sep = 10pt,
        legend columns = -1, 
        legend to name = grouplegend
    }, 
    xmode=log, 
    ymode=log,
	ylabel={Solution quality},
    title={(a) SPLP solution quality}
]
	\addplot[top conf] coordinates {
		(0.001, 0.528224027675535)
		(0.002, 0.290393145803977)
		(0.004, 0.146396592104967)
		(0.008, 0.083294403867031)
		(0.016, 0.0270881975827215)
		(0.032, 0.0148850994105934)
		(0.064, 0.00597226719078368)
		(0.128, 0.00372378429767797)
		(0.256, 0.00177145164047716)
		(0.512, 0.00142228006715113)
		(1.024, 0.000100898174527227)
	};
	\addlegendentry{Top conf.\ PP}
	\addplot[cutoff 1] coordinates {
		(0.001, 0.935431148010446)
		(0.002, 0.898058772483269)
		(0.004, 0.894458366803756)
		(0.008, 0.894458366803756)
		(0.016, 0.766057434160886)
		(0.032, 0.204886098440359)
		(0.064, 0.101287400234949)
		(0.128, 0.0385895630887643)
		(0.256, 0.00801767385429572)
		(0.512, 0.0064128552689275)
		(1.024, 0.00177013160892996)
	};
	\addlegendentry{1\% cutoff line}
	\addplot[cutoff 5] coordinates {
		(0.001, 0.947764524289625)
		(0.002, 0.947764524289625)
		(0.004, 0.947764524289625)
		(0.008, 0.947764524289625)
		(0.016, 0.947764524289625)
		(0.032, 0.903921409242321)
		(0.064, 0.255970526930509)
		(0.128, 0.145936104378611)
		(0.256, 0.0686383814820361)
		(0.512, 0.032160389665744)
		(1.024, 0.0225617062997303)
	};
	\addlegendentry{5\% cutoff line}
\nextgroupplot[
    xmode=log,
    ylabel={Configuration rank, \%},
    title={(b) SPLP configuration rank}
]
    \addplot[top conf] coordinates {
		(0.001, 32.7871316897174)
		(0.002, 4.79555021046302)
		(0.004, 3.13815393866506)
		(0.008, 3.61921226698737)
		(0.016, 1.39807576668671)
		(0.032, 1.39431749849669)
		(0.064, 0.199188214070956)
		(0.128, 0.176638604930848)
		(0.256, 0.0789236319903788)
		(0.512, 0.217979555021046)
		(1.024, 0)
	};
	\addplot[cutoff 1] coordinates {
		(0.001, 87.5)
		(0.002, 72.1737823211064)
		(0.004, 71.876879134095)
		(0.008, 72.0610342754059)
		(0.016, 59.0311184606133)
		(0.032, 10.8914612146723)
		(0.064, 7.67438364401684)
		(0.128, 5.56599518941672)
		(0.256, 2.14221286831028)
		(0.512, 2.16100420926037)
		(1.024, 0.999699338544799)
	};
	\addplot[cutoff 5] coordinates {
		(0.001, 93.539536981359)
		(0.002, 93.956704750451)
		(0.004, 93.9942874323512)
		(0.008, 94.0093205051112)
		(0.016, 94.0205953096813)
		(0.032, 74.6692723992784)
		(0.064, 15.9275405892965)
		(0.128, 9.69633193024654)
		(0.256, 7.16325917017438)
		(0.512, 5.67498496692724)
		(1.024, 4.99849669272399)
	};
    
\nextgroupplot[ 
    xmode=log, 
    ymode=log,
	ylabel={Solution quality},
    title={(c) FFMSP solution quality}
]
	\addplot[top conf] coordinates {
		(0.001, 0.578910632396653)
		(0.002, 0.263501796814842)
		(0.004, 0.241789949410404)
		(0.008, 0.212415922824409)
		(0.016, 0.170798256162075)
		(0.032, 0.157970477329623)
		(0.064, 0.145608698896416)
		(0.128, 0.0845746996305584)
		(0.256, 0.0565954343417431)
		(0.512, 0.0244154252228744)
		(1.024, 0.0130976247145846)
	};
	\addplot[cutoff 1] coordinates {
		(0.001, 0.610008428263651)
		(0.002, 0.459186303034964)
		(0.004, 0.421030818124983)
		(0.008, 0.387503574090101)
		(0.016, 0.366629626220355)
		(0.032, 0.317660594780578)
		(0.064, 0.277311616286131)
		(0.128, 0.200117220838477)
		(0.256, 0.0934891884776044)
		(0.512, 0.052691023017635)
		(1.024, 0.0200292494365965)
	};
	\addplot[cutoff 5] coordinates {
		(0.001, 0.875122575081344)
		(0.002, 0.705498044896164)
		(0.004, 0.69199016699753)
		(0.008, 0.594447600366276)
		(0.016, 0.555225861812094)
		(0.032, 0.486131745830953)
		(0.064, 0.388231503215698)
		(0.128, 0.240251845369944)
		(0.256, 0.143751517937197)
		(0.512, 0.0699681262573943)
		(1.024, 0.0256697497965267)
	};
\nextgroupplot[
    xmode=log,
    ylabel={Configuration rank, \%},
    title={(d) FFMSP configuration rank}
]
	\addplot[top conf] coordinates {
		(0.001, 51.6369047619048)
		(0.002, 31.6840277777778)
		(0.004, 40.3893849206349)
		(0.008, 46.515376984127)
		(0.016, 41.8154761904762)
		(0.032, 46.6765873015873)
		(0.064, 45.2876984126984)
		(0.128, 14.8313492063492)
		(0.256, 8.77976190476191)
		(0.512, 0.0868055555555556)
		(1.024, 0)
	};
	\addplot[cutoff 1] coordinates {
		(0.001, 89.7941468253968)
		(0.002, 80.4191468253968)
		(0.004, 76.4012896825397)
		(0.008, 91.0218253968254)
		(0.016, 93.8740079365079)
		(0.032, 93.0183531746032)
		(0.064, 89.3725198412698)
		(0.128, 77.4925595238095)
		(0.256, 36.3839285714286)
		(0.512, 16.046626984127)
		(1.024, 0.992063492063492)
	};
	\addplot[cutoff 5] coordinates {
		(0.001, 95.4613095238095)
		(0.002, 95.9325396825397)
		(0.004, 97.2842261904762)
		(0.008, 97.0610119047619)
		(0.016, 97.2470238095238)
		(0.032, 96.8377976190476)
		(0.064, 95.4861111111111)
		(0.128, 85.6522817460317)
		(0.256, 66.6914682539683)
		(0.512, 27.9885912698413)
		(1.024, 4.99751984126984)
	};
    \end{groupplot}
    \node at (4.8cm,-6.1cm) {\ref{grouplegend}}; 
\end{tikzpicture}
\negspace
\caption{Solution quality and configuration rank as they change throughout the run.}
\label{fig:splp}
\negspace
\negspace
\end{figure}
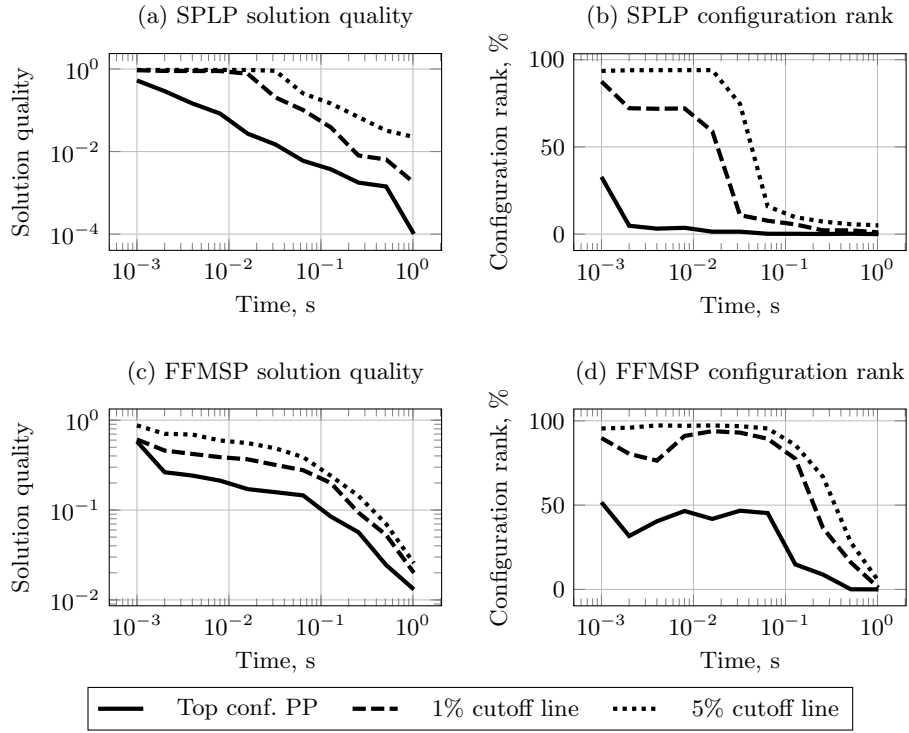

 In results for SPLP in Figure~\ref{fig:splp}(a), the solid line shows the Performance Profile (PP) of the top configuration, i.e.\ how the solution quality of the best-in-L-space configuration improves over time.
 The other two lines, which we call \emph{cutoff lines}, are from aggregating PPs of a set of configurations.
 To obtain an $x$\% cutoff line, we select the top-$x$\% of the configurations in the S-space and then produce the cutoff line as a combined worst-case PP for all those configurations.
 (All three lines are monotonic, as CMCS records the best-so-far solutions.)
 As hoped, there is a strong correlation between the short- and long-run performance of the top configurations; specifically, the cutoff lines drop relatively quickly, suggesting that there is a potential for a significant speed up by terminating configurations that perform poorly in the `long' runs.
%
 To evaluate this, we plotted the ranks corresponding to the lines in Figure~\ref{fig:splp}(b).
 The lowness of the solid line indicates the top configuration was among the top performers throughout the run.
 The drop in the cutoff lines indicates that the runs longer than ${\sim{}20}$ms are likely to be useful in early prediction of the `long' performance.
 E.g., at 32ms, the 1\% cutoff line potentially allows us to rule out 89\% of all the configurations.
 Short-run performances do link to long-runs; hence, we can use this to quickly terminate configurations that are not likely to have good L-space performance. 
%
 However, Figures~\ref{fig:splp}(c,d), showing the results for the FFMSP domain, demonstrate that in other domains one may need longer short-runs to predict `long' performance.

 Finding the exact cutoff lines in advance is impractical, as it requires long runs for all the configurations.
 The heuristic below
 obtains and exploits a reasonably reliable 1\% cutoff line much quicker.
 Specific parameter values used are given, but are changeable.
\negspace
\begin{enumerate}
    
	\item
    Randomly select 1\% out of all the configurations and place them into a pool $P$ and run full-time tests for each $c \in P$\@. 
    Generate the cutoff line as the combined worst case PP for all $c \in P$.
    
    \item 
    First pass:
    For each previously untested configuration $c$, start a full-time test and gradually build its PP\@.
    If at any of 1ms, 2ms, 4ms, \ldots, 512ms its PP rises above the cutoff line by more than 20\%, terminate the run.
    However, if $c$ survives early termination, add $c$ to $P$ and remove the worst-performing (in the L-space) configuration from $P$; update the cutoff line.
    
\ignore{    \item
    Second pass: repeat Step~2, scanning through all the early-terminated  configurations \AJP{ that were terminated early} again, but starting with the cutoff line and $P$ obtained in the first pass.
%
	\DK{Rerun only those tests that were pruned prematurely according to the new cutoff line.}
	Reuse previous results to determine configurations that had been pruned prematurely and rerun them. \AJP{previous not needed?}
	(This reduces excessive bias towards good initial performance.)
	(This recovers from potential excessive bias towards early termination in the first pass because the initial cutoff was too tight.)
}

    \item
    Second pass: repeat Step~2, but scanning only through the configurations that were terminated early. 
    This pass helps recovery from the potential bias at the beginning of the first pass if the initial cutoffs were too tight.

    
    
    
    
    \item
    Return $P$ as an approximation of the top 1\% of all the configurations.
\end{enumerate}

\begin{table}[tb]
\begin{center}
\begin{tabular}{l@{\quad}r@{\quad}r@{\quad}r@{\quad}r}
\toprule
Domain & \#conf. & Speed up & Overlap (top conf.) & Overlap (1\%) \\
\midrule
SPLP & 26,608 & 9.4 & 100\% & 99.3\% \\
FFBSP & 8,064 & 2.1 & 100\% & 99.9\% \\
BBQP & 9,860 & 9.2 & 100\% & 98.7\% \\
\bottomrule
\end{tabular}
\end{center}
\caption{Accuracy of the cutoff approximation algorithm.}
\label{tab:heuristic}
\negspace\negspace\negspace\negspace
\end{table}


 Table~\ref{tab:heuristic} gives experimental results.
 (We did not report the results for BBQP in Figure~\ref{fig:splp} due to lack of space).
 The `Speed up' column tells how much quicker the heuristic is compared to evaluating all the configurations in S-space.
 The `Overlap (top conf.)' tells how often the heuristic finds the best performing configuration (in 100 experiments), and the `Overlap (1\%)' column tells the average overlap between the true top 1\% configurations and the ones found by our heuristic.
 Note the speed up factor significantly depends on the domain; the SPLP and BBQP domains gave more than 9x speed up, though in FFMSP, the gain is much more modest. 
 In all the domains, our heuristic procedure successfully found the best performing configuration every time, and was 99\% accurate in finding the top-1\% of configurations.

\negspace
\negspace

\section{Conclusions and Future Work}
\negspace

 The PPs arising from CMCS were shown to have sufficient structure in their behaviours such that a ``performance envelope/cut-off'' could be constructed to effectively determine when terminating a test run was safe, in that most of the good configurations would be found. 
 Behaviours of the PPs differed between domains, suggesting that dynamic adaptive methods are needed.
%
%
 We gave a heuristic method for this, that automatically strengthens the cut-offs as more PPs are collected, and so reduces overall runtime.
 

 The work here only used a PP of a simple linear aggregate over a set of different test instances.  
 However, future extensions should consider ``Performance Trajectories''; using the entire, time-dependent vector of the performances over the set of test instances.
 Initial explorations have taken such performance vectors at a given ``short time point'' and then considered them as feature vectors with labels given by the ultimate aggregate quality with a long runtime. 
 For SPLP, standard classification methods did identify the regions in the S-vector space that lead to longer term good performance.  
 This suggests that information in performance trajectories is also available that can be extracted by machine learning in order to classify (or cluster) behaviours and so potentially be used to optimise policies for when tests on configurations can be terminated.

\negspace

\bibliographystyle{splncs03}
\bibliography{cmcs}

\end{document}